%% file: arxiv2014.tex
\documentclass[11pt]{article}
\usepackage{acl2014}
\usepackage{graphicx}
\usepackage{times}
\usepackage{url}
\usepackage{latexsym}
\usepackage{amsmath}
\usepackage{minibox}
\usepackage{multirow}
\usepackage{url}
\usepackage{xcolor}

\title{Context-Dependent Fine-Grained Entity Type Tagging}

\author{Dan Gillick, Nevena Lazic, Kuzman Ganchev, Jesse Kirchner, David Huynh \\
  Google Inc. \\}

\date{}

\begin{document}
\maketitle
\begin{abstract}
Entity type tagging is the task of assigning category labels to each mention of an entity in a document. While standard systems focus on a small set of types, recent work \cite{ling2012fine} suggests that using a large \emph{fine-grained} label set can lead to dramatic improvements in downstream tasks. In the absence of labeled training data, existing fine-grained tagging systems obtain examples automatically, using resolved entities and their types extracted from a knowledge base. However, since the appropriate type often depends on context (e.g. Washington could be tagged either as \emph{city} or \emph{government}), 
this procedure can result in spurious labels, leading to poorer generalization.

We propose the task of \textit{context-dependent fine type tagging}, where the set of acceptable labels for a mention is restricted to only those deducible from the local context (e.g. sentence or document). We introduce new  resources for this task: 12,017 mentions annotated with their context-dependent fine types, and we provide baseline experimental results on this data.
\end{abstract}

\section{Introduction}

The standard entity type tagging task involves identifying entity mentions (such as \textit{Barack Obama}, \textit{president}, or \textit{he}) in natural language text, and classifying them into predefined categories like \emph{person}, \emph{location}, or \emph{organization}. 
Type tagging is useful in a variety of related natural language tasks like
coreference resolution and relation extraction, as well as for downstream
processing like question answering~\cite{lin2012noun}.
Most tagging systems only consider a small set of 3-18 type labels \cite{hirschman1997muc,tjong2003introduction,doddington2004automatic}. However, recent work by Ling and Weld \cite{ling2012fine} suggests that using a much larger set of \emph{fine-grained} types can lead to substantial improvements in relation extraction and possibly other tasks.

There exists no labeled training dataset for the fine-grained tagging task. Current systems use labels derived from knowledge bases; for example, FIGER \cite{ling2012fine} uses 112 Freebase types, and HYENA \newcite{YBHS12} uses 505 YAGO types, which are Wikipedia categories mapped to WordNet synsets. In both cases, the training and evaluation examples are obtained automatically from entities resolved in Wikipedia. The resolved entities are first assigned Freebase or Wikipedia types, and these are mapped to the final set of labels.

  One issue that remains unaddressed in using distant supervision to obtain labeled examples for fine type tagging is label noise. Most resolved entities have multiple type labels; however, not all of these labels typically apply in the context of a given document. 
For example, the entity Clint Eastwood has 30 labels in YAGO, including \textit{actor}, \textit{medalist}, \textit{entertainer}, \textit{mayor}, \textit{film director}, \textit{composer}, and \textit{defender}. 
Arguably, only a few of these labels are deducible from a given news article about Clint Eastwood; similarly, only a few types can be considered ``common knowledge'' and thus inferrable from the mention text itself. 
For applications such as database completion, finding as many correct types as possible might be appropriate, and this task has been proposed in prior work.
For other applications, such as information retrieval, it is more appropriate to find only the types evoked by a mention in its context.  
For example, a user might want to find news articles about Clint Eastwood as mayor, and every mention which talks about Clint Eastwood as an entertainer should be considered wrong in this context.

The focus of our work is \textit{context-dependent fine-type tagging}, where the set of acceptable labels for a mention is constrained to only those that are relevant to the local context (e.g. sentence, paragraph, or document). 
Similarly to existing systems, our label set is derived from Freebase, 
and training data is generated automatically from resolved entities.  
However, unlike existing systems, we address the presence of spurious labels in the training data by applying a number label-pruning heuristics to  the training examples. These heuristics demonstrably improve performance on manually annotated data.

We have prepared extensive new resources related to the context-dependent fine type tagging task. 
This includes 12,017 manually annotated mentions\footnote{Available here: https://arxiv.org/e-print/1412.1820v2} in the OntoNotes test corpus \cite{weischedel2011ontonotes}.
We hope these resources will enable more research on the problem and provide a useful basis for experimental comparison.

The paper is organized as follows.
Section \ref{sec:types} describes the process of selecting the set of type labels (which we organize into a hierarchy), as well as manual annotation guidelines and ablation procedure.
Section \ref{sec:training} describes the distant supervision process for producing training data and the label-pruning heuristics. 
Section \ref{sec:models} describes the features we used, baseline tagging models, and inference. 
Section \ref{sec:experiments} outlines evaluation metrics and walks through a series of experiments that produced the best results.

\section{Fine-grained type labels and manual annotations}
\label{sec:types}

\input{hierarchy.tex}

Our set of fine grained type labels $\mathcal{T}$ is derived from Freebase similarly to FIGER; however, we additionally organize the labels into a hierarchy. The motivation for this is that a hierarchy allows us to incorporate simple domain knowledge (for example, that an \emph{athlete} is also a \emph{person}, but not a \emph{location}) and ensure label consistency. Furthermore, if the number of possible labels is very large, it allows for faster inference by assigning labels in a top-down manner. 

The labels are organized into a tree-structured taxonomy, where each label is related to its parent in the tree via the asymmetric, anti-reflexive, transitive ``IS-A" relationship.
The root of the tree is a default node encompassing all types. Labels at the first level of the tree are the commonly used coarse types \emph{person}, \emph{location}, \emph{organization}, and \emph{other}.
These are then subdivided into more fine-grained categories, as illustrated in Fig.~\ref{fig:taxonomy}.
The taxonomy was assembled manually using an iterative process, starting with Ling and Weld's non-hierarchical types. We organized the types into a hierarchy and then refined them, removing labels that seemed rare or ambiguous, and including new labels if there were enough examples to justify the addition.
In general, we preferred a taxonomy that would yield a single label path for each mention and have as little ambiguity as possible. 

Once this process was finalized, we mapped the common (non-hierarchical) Freebase types to our types. 
Whenever there was any ambiguity in the mapping, we backed off to a common parent in the hierarchy.

Finally, we note that the set of 505 YAGO types used in \cite{YBHS12} is also hierarchical, with 5 top-level types and 100 labels in each subcategory. Most of our types can be mapped to YAGO types in a straight-forward manner. However, we found that using the YAGO labels directly would lead to much ambiguity in manual annotations, primarily due to the large number of labels, the directed-acyclic-graph nature of the hierarchy, and the presence of `qualitative' labels (e.g. \textit{person/good person}, \emph{event/happening/beginning}).

\subsection{Manual annotations}

Type annotations are meant to be context dependent; that is, the only assigned types should be those that are deducible from the sentence, or perhaps the paragraph.
Of course, the notions of \emph{deducibility} and \emph{local context} are subjective.
The annotators were instructed to label ``San Francisco'' as a \emph{location/city} even if this is not made explicit in the context, since this can be considered common knowledge and should be inferrable from the mention text itself.
On the other hand, in the case of any uncertainty, the annotators were instructed to back off to the parent type (in this case,  \emph{location}).

The corpus we annotated for this work includes all news
documents in the OntoNotes test set except the longest 4, which we dropped to reduce annotator burden.
Table~\ref{tab:evalcorpus} summarizes some of the corpus statistics and provides example annotations. 
Note that labels at level 2 (e.g. \emph{person/artist}) are
approximately half as common as labels at level 1 (e.g. \emph{person}), but are
10 times as common as labels at level 3. 
The main reason for this is that we allow labels to be partial paths in the hierarchy tree (i.e. root to internal node, as opposed to root to leaf), and some of the level 3 labels rarely occur in the training data. Furthermore, many of the level 2 types have no sub-types; for example \emph{person/athlete} does not have separate sub-categories for swimmers and runners.

\begin{table*}[t]
  \centering
  \begin{tabular}{@{}l@{\hspace{0.4in}}r@{}}
  \begin{tabular}{|l|r|}
    \hline
    \textbf{Statistic} & \textbf{Value} \\
    \hline
    Documents & 77 \\
    Entity mentions & 12017 \\
    Labels & 17704 \\
    Labels at Level 1 & 11909 \\
    Labels at Level 2 & 5209 \\
    Labels at Level 3 & 586 \\
    \hline
  \end{tabular}
  &
  \begin{tabular}{l}
   \textbf{Text:} If a hostile \fbox{predator} emerges for 
   \fbox{Saatchi \& Saatchi Co.},  \\
   \fbox{co-founders Charles and Maurice Saatchi} will lead \ldots \\
   \\
   \textbf{Fine Types:} \\
   predator: \emph{organization/company}, \emph{other} \\
   Saatchi \& Saatchi Co.: \emph{organization/company} \\
   co-founders Charles and Maurice Saatchi: \emph{person/business} \\
  \end{tabular}
  \end{tabular}
  \caption{\label{tab:evalcorpus}
    Corpus statistics (left) and example (right). Level 1, 2 and 3 correspond
    to levels in the label
    hierarchy in Figure~\ref{fig:taxonomy}.  For example, Level 2 includes
    labels as \emph{person/artist} while Level 3 are one level lower such as
    \emph{person/artist/actor}.  Entities in the example are inside a box.
  }
\end{table*}

We built an interactive web interface for annotators to quickly apply types to mentions (including named, nominal, and pronominal mentions); on average, this task took about 10 minutes per document.
Six annotators independently labeled each document and we kept the labels with support from at least two of the annotators (about 1 of every 4 labels was pruned as a result).
It is worth distinguishing between two kinds of label disagreements.
\emph{Specificity} disagreements arise from differing interpretations of the appropriate depth for a label, like \emph{person/artist} vs. \emph{person/artist/actor}.
More problematic are \emph{type} disagreements arising from differing interpretations of a mention in context or of the type definitions.

Applying the agreement pruning reduces the total number of pairwise disagreements from 3900 to 1303 (specificity) and 3700 to 774 (type).
The most common remaining disagreements are shown in Table \ref{tab:disagreements}.
Some of these could probably be eliminated by extra documentation.
For example, in the sentence ``Olivetti has denied that it violated Cocom rules'', the mention ``rules'' is labeled as both \textit{other} and \textit{other/legal}.
While it is clear from context that this is indeed a legal issue, the examples provided in the annotation guidelines are more specific to laws and courts (``5th Amendment'', ``Treaty of Versailles'', ``Roe v. Wade'').
In other cases, the assignment of multiple types may well be correct:
``Syrians'' in ``...whose lobbies and hallways were decorated with murals of ancient Syrians...'' is labeled with both \textit{person} and \textit{other/heritage}.

\begin{table}[htb]
  \centering
  \begin{tabular}{|l|l|r|}
    \hline
    \textbf{Label 1} & \textbf{Label 2} & \textbf{Count} \\
    \hline
    \textit{Other} & \textit{Other/legal} & 227 \\
    \textit{Other} & \textit{Other/product} & 155 \\
    \textit{Person} & \textit{Person/business} & 125 \\
    \textit{Other} & \textit{Other/currency} & 90 \\
    \textit{Person} & \textit{Person/political-figure} & 89 \\
    \hline
    \hline
    \textit{Other} & \textit{Organization/company} & 126 \\
    \textit{Other} & \textit{Person} & 101 \\
    \textit{Other} & \textit{Location} & 90 \\
    \textit{Other} & \textit{Organization} & 50 \\
    \textit{Person/title} & \textit{Person/business} & 38 \\
    \hline
  \end{tabular}
  \caption{\label{tab:disagreements} The most common specificity disagreements (top) and type disagreements (bottom) observed in the test data after removing labels applied by fewer than two annotators.}
\end{table}

We assessed the difficulty of the annotation task using average annotator precision, recall and F1 relative to the consensus (pruned) types, shown in Table \ref{tab:agreement}. As expected, there is less agreement
over types that are deeper in the hierarchy, but the high precision (92\% at depth 2 and 89\% at depth 3) reassures us that the context-dependent annotation task is reasonably well defined.

\begin{table}[htb]
  \centering
  \begin{tabular}{|l|r|r|r|}
    \hline
    \textbf{Depth} & \textbf{Precision} & \textbf{Recall} & \textbf{F1} \\
    \hline
    1 & 0.98 & 0.93 & 0.96 \\
    2 & 0.92 & 0.76 & 0.83 \\
    3 & 0.89 & 0.69 & 0.78 \\
    \hline
  \end{tabular}
\caption{\label{tab:agreement} Average annotator precision, recall and F1 with respect to the consensus types.}
\end{table}

Finally, we compared the manual annotations to the labels obtained automatically from Freebase for the resolved entities in our data.
The overall recall was fairly high (80\%), which is unsurprising since Freebase-mapped types are typically a superset of the context-specific type. However, precision was low (50\%), suggesting that many of the automatically generated types are unrelated to mention contexts.

\section{Distant supervision for training}
\label{sec:training}

\subsection{Assembling training data}
Ling and Weld use the internal links in Wikipedia as training data:
a linked entity inherits the Freebase types associated with the landing page.
We adopt a similar strategy, but rely instead on an entity resolution system that assigns Freebase types to resolved entities, which we then map to our types.

We use a set of 133,000 news documents as the training corpus.
Each document is processed by a standard NLP pipeline.
This includes a part-of-speech (POS) tagger and dependency parser, comparable in accuracy to the current Stanford dependency parser \cite{klein2003accurate}, and an NP extractor that makes use of POS tags and dependency edges to identify a set of entity mentions.
Thus we separate the type tagging task from the identification of entity mentions, often performed jointly by entity recognition systems.
Lastly, our entity resolver links entity mentions to Freebase profiles; the system maps string aliases (``Barack Obama'', ``Obama'', ``Barack H. Obama'', etc.) to profiles with probabilities derived from Wikipedia anchors.

Next, we apply the types induced from Freebase to each entity. As already discussed, this can introduce
label noise.
For example, Barack Obama is both a \textit{person/political-figure} and a \textit{person/artist/author}, even though only one of these may be deducible from the local context of a mention.
This issue is discussed by Ritter et al. \cite{ritter2011named} in relation to entity recognition (with 10 types) for Twitter messages, and is addressed by constraining the set of types to those consistent with a topic model.
Instead, we attempt to reduce the mismatch between training and our manually-annotated test data using a set of heuristics.

\subsection{Training heuristics}
\label{sec:heuristics}

\noindent \textbf{Sibling pruning} \\
The first heuristic that we apply to refine the training data removes sibling types associated with a single entity, leaving only the parent type.
For example, an entity with types \textit{person/political-figure} and \textit{person/athlete} would end up with a single type \textit{person}. The motivation for this heuristic is that it is uncommon for several sibling types to be relevant in the same context.
This may remove some correct labels; for example, instances of Barack Obama will only be tagged with \textit{person}, even though in many cases, \textit{person/political-figure} is correct. 
However, less common entities associated with few Freebase types are better for generating training data, as they are usually annotated with types relevant to the context. 
Thus we learn about politicians from mayors and governors rather than from presidents.

\vspace{8 pt} \noindent \textbf{Coarse type pruning} \\
The second heuristic removes types that do not agree with the output of a
standard coarse-grained type classifier trained on the set of types
\{\textit{person, location, organization, other}\}. We use a softmax classifier trained on labeled data derived from ACE
\cite{doddington2004automatic}. We apply a simple label mapping to the four
coarse types, and use features similar to those described in Klein et al.
\shortcite{klein2003named}. The motivation here is to reduce ambiguity by
encouraging type labels to correspond to a single subtree of a hierarchy.
Furthermore, if the entity is annotated with conflicting types (e.g.
\emph{location} and \emph{organization}), this heuristic can help select the
type more appropriate to the context.

\vspace{8 pt} \noindent \textbf{Minimum count pruning} \\
The third heuristic removes types that appear fewer than some minimum number of times in the document (in our experiments, we require each type to appear at least twice).
The intuition is that types relevant to the document context (for example \textit{organization/sports-team} in a sports article) are likely to apply to multiple mentions in a document. 

\vspace{8 pt}
Because the heuristics prune potentially spurious labels, they decrease the total number of training examples.
Table~\ref{tab:heuristics} in the Experiments section shows the number of resulting training instances with each type of heuristic.
Finally, we note that there exist non-trivial interactions between the heuristics.  For example, Barack Obama, is associated with types \textit{person}, \textit{person/political-figure} and \textit{person/artist}, and the Sibling heuristic would normally prune these to \textit{person}. 
However, if another heuristic prunes out \textit{person/artist}, then the input to the Sibling heuristic would be just \textit{person} and \textit{person/artist}, resulting in no additional pruning.
The heuristics are applied in the order in which they are introduced above.

\section{Feature extraction, models, and inference}
\label{sec:models}

\subsection{Feature extraction}

For each mention of a resolved entity with at least one type, we extract a training instance $({\bf x}, {\bf y})$, where ${\bf x}$ is a vector of binary feature indicators and ${\bf y} \in \{0, 1\}^{|\mathcal{T}|}$ is the binary vector of label indicators.
The feature set includes the lexical and syntactic features described in Table \ref{tab:features}, similar to those used in previous work. We also use a more semantic document topic feature, the result of training a simple bag-of-words topic model with eight topics (arts, business, entertainment, health, mayhem, politics, scitech, sport), to try to capture longer-range context.
The word clusters are derived from the class-based exchange clustering algorithm described by Uszkoreit and Brants \shortcite{uszkoreit2008distributed}.

\begin{table*}[t!]
  \centering
  \begin{tabular}{|l|l|l|}
    \hline
    \textbf{Feature} & \textbf{Description} & \textbf{Example} \\
    \hline
    Head & The syntactic head of the mention phrase & ``Obama'' \\
    Non-head & Each non-head word in the mention phrase & ``Barack'', ``H.'' \\
    Cluster & Word cluster id for the head word & ``59'' \\
    Characters & Each character trigram in the mention head & ``:ob'', ``oba'', ``bam'', ``ama'', ``ma:'' \\
    Shape & The word shape of the words in the mention phrase & ``Aa A. Aa'' \\
    Role & Dependency label on the mention head & ``nsubj'' \\
    Context & Words before and after the mention phrase & ``B:who'', ``A:first'' \\
    Parent & The head's lexical parent in the dependency tree & ``picked'' \\
    Topic & The most likely topic label for the document & ``politics'' \\
    \hline
  \end{tabular}
  \caption{\label{tab:features} List of features used in type tagging. Features are extracted from each mention. Context used for example features: ``... who [Barack H. Obama] first picked ...''}
\end{table*}

Intuitively, the features describing the mention phrase itself are most relevant for the top level of the type taxonomy, while distinguishing types deeper in the taxonomy requires more contextual features.
We use the same feature representation for all types; the relevant features for each type get weighted appropriately during learning.
However, it may be worthwhile to make this distinction explicit in future work, and the hierarchy levels are a convenient structure for applying different feature sets.

\subsection{Models and inference}

Hierarchical classification can be seen as a special case of structured multilabel classification, where the output space is a class taxonomy. A recent survey \cite{Silla11} categorizes existing approaches as:
\textit{flat}, using a single multiclass classifier, \textit{local}, using a binary classifier for each label and enforcing label consistency at test time, \textit{local per parent node}, using a multiclass classifier for all children of a node, and \textit{global}, training a single multiclass classifier but replacing the standard zero-one loss with a function that reflects label similarity. 

We explore the baseline flat and local approaches, and acknowledge that results can possibly be improved using more complex models. In particular, we use the maximum entropy discriminative local and flat classifiers (i.e. logistic and softmax regression). We note that existing fine-type tagging systems also rely on simple linear classifiers; FIGER uses a flat multi-class perceptron, allowing multiple labels as output, while HYENA employs multiple binary support vector machine (SVM) classifiers with some postprocessing of the outputs.  In general, the discriminative ability of any classifier diminishes as the number of classes increases, so we expect local classifiers to outperform a flat one. This is confirmed empirically in our experiments, as well as in existing work (i.e. HYENA outperforms FIGER).

\subsubsection{Local classifiers}
\label{sec:local}

In the local approach, a binary classifier is independently trained for each label, and label consistency is enforced at inference time. For each label $t$, we train a binary logistic regression classifier with $L2$ regularization.

Defining the positive and negative training examples for each binary classifier is not entirely straightforward, due to the asymmetric IS-A relationships between the labels. We set the positive examples for a type to itself and all its descendants in the hierarchy; for example, a mention labeled $\emph{person/artist}$ is considered a positive example for $\emph{person}$.
We experiment with setting the negative examples for a type as (1) all other types with the same parent, (2) all other types at the same depth, or (3) all other types.

At inference time, given the learned parameters and a test feature vector ${\bf x}$, we first independently evaluate the probability of each type. We then consider the following three inference strategies for assigning labels:

\vspace{8 pt} \noindent $\bullet$ \emph{Independent}. We assign all types whose probability exceeds some decision threshold, without enforcing the labels to correspond to a single path in a hierarchy.

\vspace{8 pt} \noindent $\bullet$ \emph{Conditional}. We multiply the probability of each label $t$ by the probability of its parent $pa(t)$ for all types other than the top-level coarse types. This strategy ensures that if a label $t$ is assigned at a given decision threshold, $pa(t)$ must be assigned as well; however, it does allow for multiple paths in the hierarchy tree.

\vspace{8 pt} \noindent $\bullet$ \emph{Marginalizing out IS-A constraints}. We refine the probability of each label by
marginalizing out the hierarchy constraints. Specifically, we first compute the probability of each valid label configuration (each path from root to a leaf or internal node in the hierarchy) as
\begin{equation}
p({\bf y}) \propto 
\begin{cases}
\prod_t p(y_t) & \text{${\bf y}$ is a path} \\
0 & \text{otherwise.}
\end{cases}
\end{equation}
We then set the probability of an individual label $t$ to the sum of the probabilities of configurations in which $y_t=1$. Since the number of paths is not too large, we simply list all paths; with larger label sets, the marginalization can be done more efficiently using the sum-product algorithm. We assign all labels whose refined probabilities are above a given threshold.

\subsubsection{Flat classifier}
\label{sec:flat}

In this approach, we train a flat softmax regression classifier~\cite{berger1996maximum} to discriminate
between all possible types. This classifier expects a single type label to each
instance, whereas our training examples are labeled with multiple types. To
account for this, at training time, we convert each multi-label instance to multiple
single-label instances. For example, an occurrence of ``Canada'' could be both
\emph{location} and \emph{organization}.  Rather than constructing a learning
objective appropriate for such multi-label training data, we produce two
training examples, one with label \emph{location} and the other with label
\emph{organization}. 
At inference time, we assign all labels whose probability exceeds a threshold,
rather than selecting a single highest scoring label. 

\begin{table*}[ht!]
\centering
\begin{minipage}[b]{0.45\linewidth}
  \centering
  \begin{tabular}{|l|rrrr|}
    \hline
    \textbf{Negatives} & \textbf{Prec} & \textbf{Rec} & \textbf{F1} & \textbf{AUC} \\
    \hline
    All     &    77.98 &    59.55 &    67.53 &    66.56 \\
    Sibling &    79.93 &    58.94 &    67.85 &    66.50 \\
    Depth   &\bf 80.05 &\bf 62.20 &\bf 70.01 &\bf 69.29 \\
    \hline
  \end{tabular}
\end{minipage}
\hspace{0.5cm}
\begin{minipage}[b]{0.45\linewidth}
  \centering
  \begin{tabular}{|l|rrrr|}
    \hline
    \textbf{Inference} & \textbf{Prec} & \textbf{Rec} & \textbf{F1} & \textbf{AUC} \\
    \hline
    Independent &    77.06 &    61.54 &    68.43 &    67.74 \\
    Conditional &    77.89 &\bf 63.30 &    69.84 &\bf 70.04 \\
    Marginals   &\bf 80.05 &    62.20 &\bf 70.01 &    69.29 \\
    \hline
\end{tabular}
\end{minipage}
  \caption{\label{tab:training} A comparison of methods for choosing negative
    examples for local classification (left) and inference schemes (right).
    In the experiments on the left, we perform inference by marginalizing out constraints,
    and in the experiments on the right, we
    use Depth to choose negative examples; both use all pruning heuristics.
    The best scores are in bold.
  }
\end{table*}

\section{Experiments}
\label{sec:experiments}

Assessing the performance of a hierarchical classifier is not straightforward.  
Previous work introduces a variety of loss measures to evaluate
hierarchical classification errors; see for
example~\newcite{cesa2006incremental} or~\newcite{weinberger2008large}.  For
simplicity, we evaluate performance using Precision, Recall, F-score and area under the
precision/recall curve.  Since performance metrics are dominated by the level 1 types, we additionally report  
precision, recall, and F-score at each level (see Table \ref{tab:classifiers}).

We split the gold data into a development set with 16 documents, and a test set with 61 documents, and report results on the test set.
We only evaluate named and nominal mentions (11197 non-proniminal mentions), as is standard in the named entity recognition literature.
For the sake of simplicity, we choose a single threshold that maximizes overall F-score on the development set. We do observe a wide range of precision/recall numbers for the individual labels, so using label-specific thresholds might give better results.

\subsection{Classifiers and inference}

We start by evaluating the local classifier approach described in Section \ref{sec:local}. We compare the three  strategies for selecting negative examples, as well as the three inference methods for assigning labels. For each training strategy, we report the results of the best corresponding inference method, and vice versa. The results are presented in Table \ref{tab:training}; the best results were obtained using same-depth labels as negative training examples, and marginalizing out hierarchy constraints.

Next, we compare the best local classifier results to the flat classifier described in Section \ref{sec:flat}.
Note that the features and the total number of model parameters are identical for the two approaches.
The results are presented in Table \ref{tab:classifiers} and indicate that the local classifier outperforms the flat classifier, especially at deeper levels.
The area under the precision/recall curve (AUC) is 63.7\% for the flat classifier and 69.3\% for the local classifier.

\begin{table}[t!]
  \centering
  \begin{tabular}{|l|rrr|}
    \hline
    \textbf{Classifier} & \textbf{Precision} & \textbf{Recall} & \textbf{F1} \\
    \hline
    Level 1 Flat  &    84.39 &    79.01 &    81.61 \\
    Level 1 Local &    87.12 &    78.84 &\bf 82.80 \\
    \hline
    Level 2 Flat  &    46.61 &    25.99 &    33.37 \\ 
    Level 2 Local &    56.76 &    30.88 &\bf 40.00 \\ 
    \hline
    Level 3 Flat  &    75.00 &    1.78  &     3.47 \\ 
    Level 3 Local &    24.00 &    8.28  &\bf 12.32 \\ 
    \hline
  \end{tabular}
  \caption{\label{tab:classifiers} Precision, recall, and F-Score given by the
  flat and local classifiers at each level of the type taxonomy.  We use all
  heuristics and Depth negative examples for the local classifiers. Level 1 are
  the labels immediately below the root of our tree: \emph{person},
  \emph{location}, \emph{organization}, and \emph{other}. Level 2 are the
  labels below them such as \emph{person/artist} while Level 3 are one level
  lower such as \emph{person/artist/actor}.}
\end{table}

\subsection{Distant supervision heuristics}
\label{sec:exp_heuristics}

We compare the effects of different heuristics for pruning training labels in
Table \ref{tab:heuristics}, with the best settings for our models: using local
classifiers with same-depth negative examples and marginalizing over constraints
at inference time.
Table~\ref{tab:heuristics} lists also the number of training examples extracted
from the data, as discussed in Section~\ref{sec:heuristics}.  It is evident that
the heuristics have a significant effect on system performance, with the coarse
pruning being particularly important. Together, the heuristics improve overall
F1 by 11.3\% and the AUC by 7.2\%. 

\begin{table*}[htb]
  \centering
  \begin{tabular}{cc}
  \hspace*{-0.8in}
  \begin{tabular}{c}
  \vspace{-0.2in}
  \includegraphics[width=2.8in]{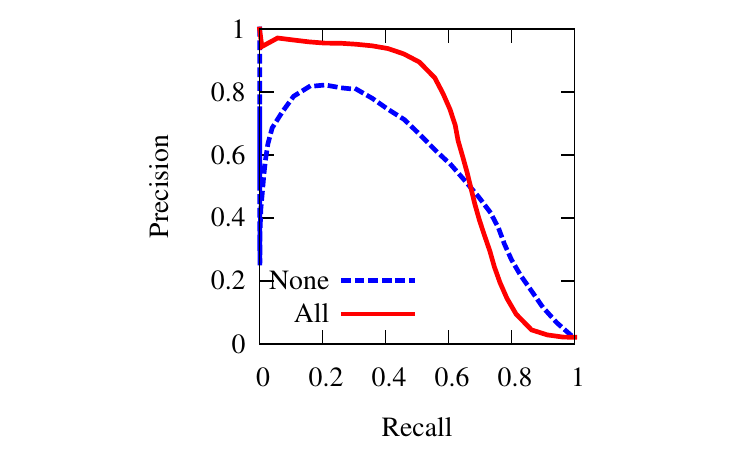}
  \hspace*{-0.5in}
  \end{tabular}
  \vspace{0.1in}
  \begin{tabular}{|l|rrrrr|}
    \hline
    \textbf{Heuristic} & \textbf{Examples} & \textbf{Precision} & \textbf{Recall} & \textbf{F1} & \textbf{AUC} \\
 & \textbf{(millions)} & &  &  &  \\
    \hline
    None        &  8.58 & 67.19 & 57.56 & 62.00 & 63.62 \\
    Min-Count=2 &  8.15 & 67.96 & 59.05 & 63.20 & 64.87 \\
    Sibling     &  3.07 & 74.12 & 57.23 & 64.59 & 66.44 \\
    Coarse      &  6.45 & 73.21 & 59.38 & 65.57 & 67.36 \\
    All         &  5.08 & 80.05 & 62.20 & 70.01 & 69.29 \\
    \hline
  \end{tabular}
  \end{tabular}
  \caption{\label{tab:heuristics} A comparison of the effects of label pruning heuristics on the system performance. Examples refers to the total number of training examples extracted from the data. Each heuristic alone improves on the baseline, and together, the improvement is largest, particularly in precision. }
\end{table*}

\section{Discussion and conclusions}

Entity type tagging is a key component of many natural language systems such as relation extraction and coreference resolution. Evidence suggests that performance of such systems can be dramatically improved by using fine-grained type tags in place of the standard limited set of coarse tags. In the absence of labeled training data, fine type tagging systems typically obtain training data automatically, using resolved entities and types extracted from a knowledge base. As entities often have multiple assigned types, this process can result in spurious type labels, which are neither obvious from the local context, nor considered common knowledge.  This subtle issue is not addressed in existing systems, which are both trained and evaluated on automatically generated data. 

In this paper, we strive to make fine-type tagging more meaningful by requiring context-dependence; that is, we require the assigned labels to be deducible from local context. To this end, we introduce several distant supervision heuristics that are aimed at pruning irrelevant labels from the training data. The heuristics reduce the mismatch between the training and gold data, and lead to a significant improvement in performance. Finally, in order to provide a meaningful basis for experimental comparison, we introduce new resources for the task, including 12,017 manually-annotated mentions in 77 OntoNotes news documents with 17,704 type labels. 

Our experimental results highlight some of the difficulties in performing type tagging with a large label set, especially in the case of very specific labels for which there are relatively few examples. There exist many directions for future work in this area. For example, we could consider jointly labeling multiple mentions within the same document, since their labels are likely correlated and some may be coreferent. In our current system, such correlations are only handled implicitly, through the document topic feature.  Since our labels are organized in a natural hierarchy, it is also worth considering richer models designed specifically for hierarchical classification problems. Finally, we can consider adding more specific label constraints in addition to those imposed by the hierarchy; for example, we might allow some multi-path labels (e.g. \emph{location}, \emph{organization}), but not others (e.g. \emph{person}, \emph{location}). 

We believe that our problem formulation, training heuristics, and new resources will help provide a meaningful framework for future research on this problem.

\bibliographystyle{acl}

\bibliography{arxiv2014}

\end{document}

%% file: hierarchy.tex
\def\dx{0.16cm}
\def\ddx{0.08cm}
\def\dy{8cm}

\def\maincolor{gray!10}
\def\subcolor{pink!10}

\begin{figure*}[t!]
\small
\raggedright
\fcolorbox{black}{\maincolor}{
\begin{minipage}[b][\dy][t]{0.17\linewidth}
\hspace*{0.5cm}{\bf \textsc{PERSON}} \\ \\
\fcolorbox{black}{\subcolor}{
\parbox{0.8\textwidth}{
{\bf artist}\\
{
\hspace*{\dx} actor \\
\hspace*{\dx} author\\
\hspace*{\dx} director\\
\hspace*{\dx} music }}} \\
\fcolorbox{black}{\subcolor}{
\parbox{0.8\textwidth}{
{\bf education}\\
{
\hspace*{\dx} student \\
\hspace*{\dx} teacher }}} \\
{\bf \hspace*{\ddx}athlete \\ 
\hspace*{\ddx}business \\ 
\hspace*{\ddx}coach\\
\hspace*{\ddx}doctor \\ 
\hspace*{\ddx}legal \\
\hspace*{\ddx}military \\
\hspace*{\ddx}political figure \\ 
\hspace*{\ddx}religious leader \\
\hspace*{\ddx}title}\\
\end{minipage}}
\fcolorbox{black}{\maincolor}{
\begin{minipage}[b][\dy][t]{0.17\linewidth}
\hspace*{0.2cm}{\bf \textsc{LOCATION}} \\ \\
\fcolorbox{black}{\subcolor}{
\parbox{0.85\textwidth}{
 {\bf structure} \\ 
{
\hspace*{\dx}airport \\ 
\hspace*{\dx}government\\ 
\hspace*{\dx}hospital\\ 
\hspace*{\dx}hotel \\ 
\hspace*{\dx}restaurant \\ 
\hspace*{\dx}sports
facility \\ 
\hspace*{\dx}theatre }}} \\
\fcolorbox{black}{\subcolor}{
\parbox{0.85\textwidth}{
 {\bf geography} \\ 
{
\hspace*{\dx}body of water \\ 
\hspace*{\dx}island\\  
\hspace*{\dx}mountain }}} \\
\fcolorbox{black}{\subcolor}{
\parbox{0.85\textwidth}{
 {\bf transit} \\ 
{
\hspace*{\dx}bridge \\ 
\hspace*{\dx}railway\\  
\hspace*{\dx}road }}} \\
{\bf 
\hspace*{\ddx}celestial \\ 
\hspace*{\ddx}city \\ 
\hspace*{\ddx}country \\ 
\hspace*{\ddx}park}
\end{minipage}}
\fcolorbox{black}{\maincolor}{
\begin{minipage}[b][\dy][t]{0.2\linewidth}
\hspace*{0.2cm}{\bf \textsc{ORGANIZATION}} \\ \\
\fcolorbox{black}{\subcolor}{
\parbox{0.85\textwidth}{ 
{\bf company} \\ 
{
\hspace*{\dx}broadcast \\
\hspace*{\dx}news }}}\\
{\bf \hspace*{\ddx}education \\
\hspace*{\ddx}government\\
\hspace*{\ddx}military \\
\hspace*{\ddx}music \\
\hspace*{\ddx}political party \\
\hspace*{\ddx}sports league \\
\hspace*{\ddx}sports team \\
\hspace*{\ddx}stock exchange \\
\hspace*{\ddx}transit}
\end{minipage}}
\fcolorbox{black}{\maincolor}{
\begin{minipage}[b][\dy][t]{0.38\linewidth}
\hspace*{2cm} {\bf \textsc{OTHER}} \\ \\
\begin{minipage}[b][6.5cm][t]{0.48\linewidth} 
\fcolorbox{black}{\subcolor}{
\parbox{0.9\textwidth}{ 
{\bf art} \\ 
{
\hspace*{\dx}broadcast \\
\hspace*{\dx}film \\
\hspace*{\dx}music\\
\hspace*{\dx}stage\\
\hspace*{\dx}writing}}}\\
\fcolorbox{black}{\subcolor}{
\parbox{0.9\textwidth}{ 
{\bf event} \\ 
{
\hspace*{\dx}accident \\ 
\hspace*{\dx}election \\ 
\hspace*{\dx}holiday\\ 
\hspace*{\dx}natural 
disaster\\
\hspace*{\dx}protest\\
\hspace*{\dx}sports event \\ 
\hspace*{\dx}violent conflict }}}\\
\fcolorbox{black}{\subcolor}{
\parbox{0.9\textwidth}{ 
 {\bf health} \\ 
{
\hspace*{\dx}malady \\ 
\hspace*{\dx}treatment }}}\\
{\bf 
\hspace*{\ddx}award \\ 
\hspace*{\ddx}body part \\ 
\hspace*{\ddx}currency   } \\
\end{minipage} \hfill
\begin{minipage}[b][6.5cm][t]{0.48\linewidth} 
\fcolorbox{black}{\subcolor}{
\parbox{0.8\textwidth}{ 
{\bf language} \\
{\hspace*{\dx}programming}\\
{ language }}}\\
\fcolorbox{black}{\subcolor}{
\parbox{0.8\textwidth}{ 
{\bf living thing} \\
{
 \hspace*{\dx}animal }}}\\ 
\fcolorbox{black}{\subcolor}{
\parbox{0.8\textwidth}{ 
{\bf product} \\ 
{\hspace*{\dx}camera\\ 
\hspace*{\dx}car\\ 
\hspace*{\dx}computer\\
\hspace*{\dx}mobile phone\\
\hspace*{\dx}software
\hspace*{\dx}weapon}}}\\
{\bf
\hspace*{\ddx}food \\
\hspace*{\ddx}heritage \\
\hspace*{\ddx}internet\\
\hspace*{\ddx}legal\\
\hspace*{\ddx}religion \\
\hspace*{\ddx}scientific \\  
\hspace*{\ddx}sports \& leisure \\ 
\hspace*{\ddx}supernatural }
\end{minipage}
\end{minipage}}
\label{fig:taxonomy}
\caption{Our type taxonomy includes types at three levels, e.g. {\textsc{PERSON}} (level 1), {\bf artist} (level 2), {actor} (level 3). Each assigned type (such as {\bf artist}) also implies the more general ancestor types (such as \textsc{PERSON}). The top level types were chosen to align with the most common type set used in traditional entity tagging systems.}
\end{figure*}